\documentclass[11pt,review]{elsarticle_1}

\usepackage{geometry}
\usepackage{fancyhdr}
\usepackage{enumerate}
\usepackage{graphicx,floatrow}
\usepackage{subfigure}
\usepackage{graphicx}
\usepackage{amsmath,amssymb}
\usepackage{url}
\usepackage{float}
\usepackage{color}
\usepackage{multirow}
\usepackage[mathscr]{euscript}
\usepackage[ruled]{algorithm2e}

\renewcommand{\algorithmcfname}{ALGORITHM}
\SetAlFnt{\small}
\SetAlCapFnt{\small}
\SetAlCapNameFnt{\small}
\SetAlCapHSkip{0pt}
\IncMargin{-\parindent}

\begin{document}


\title{Video Face Editing Using Temporal-Spatial-Smooth Warping}


\author[]{Xiaoyan Li}
\author[]{Dacheng Tao}
\address{University of Technology, Sydney}

\begin{abstract}
Editing faces in videos is a popular yet challenging aspect of computer vision and graphics, which encompasses several applications including facial attractiveness enhancement, makeup transfer, face replacement, and expression manipulation. Simply applying image-based warping algorithms to video-based face editing produces temporal incoherence in the synthesized videos because it is impossible to consistently localize facial features in two frames representing two different faces in two different videos (or even two consecutive frames representing the same face in one video). Therefore, high performance face editing usually requires significant manual manipulation. In this paper we propose a novel temporal-spatial-smooth warping (TSSW) algorithm to effectively exploit the temporal information in two consecutive frames, as well as the spatial smoothness within each frame. TSSW precisely estimates two control lattices in the horizontal and vertical directions respectively from the corresponding control lattices in the previous frame, by minimizing a novel energy function that unifies a data-driven term, a smoothness term, and feature point constraints. Corresponding warping surfaces then precisely map source frames to the target frames. Experimental testing on facial attractiveness enhancement, makeup transfer, face replacement, and expression manipulation demonstrates that the proposed approaches can effectively preserve spatial smoothness and temporal coherence in editing facial geometry, skin detail, identity, and expression, which outperform the existing face editing methods. In particular, TSSW is robust to subtly inaccurate localization of feature points and is a vast improvement over image-based warping methods.
\end{abstract}

%
%




%

\maketitle

\section{Introduction}
\label{sec:introd}
The dramatic growth in the availability of online videos has resulted in a greater demand for editing the faces that appear in videos. In practice, the four most required video editing applications are: 1) enhancing facial ``attractiveness'' in synthesized videos; 2) transferring makeup from one face to another face; 3) replacing the face in the target video with a source face; and 4) manipulating (e.g., exaggerating or neutralizing) facial expressions while preserving facial identity.

\begin{figure}[!t]
\normalsize
\centering
\includegraphics[width=13.5cm]{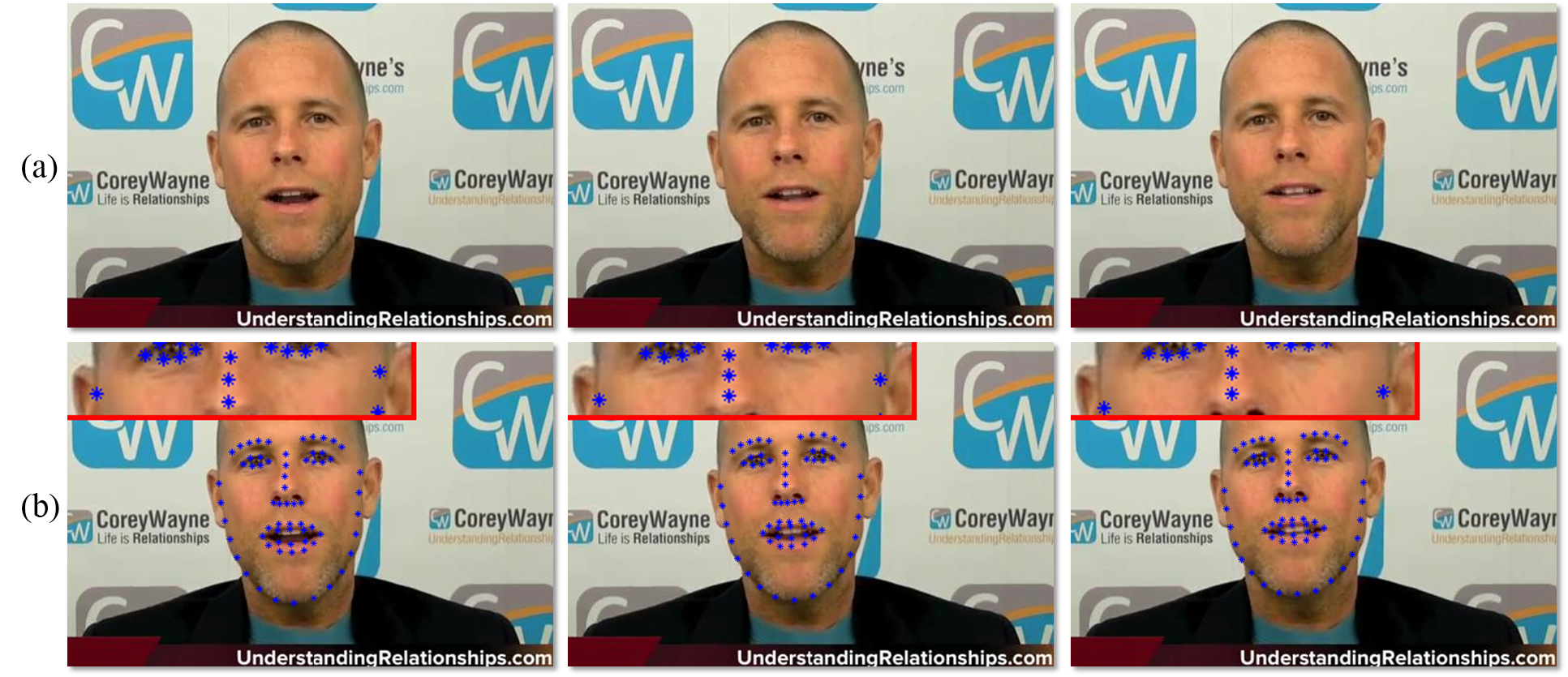}
\caption{Inconsistent localization. (a) Three successive frames in an original video. (b) Corresponding facial feature point localization using an automatic tracking technique. Subtly inconsisten localization often occurred between consecutive frames (see 2$\times$ local magnification in the top-left corner).}
\label{fig:inconsist}
\end{figure}

Achieving these aims is not as simple as directly applying an image-based face editing technique (e.g., \cite{Leyvand:2008}, \cite{Tong:2007}, \cite{Guo:2009}, \cite{Scherbaum:2011:Makeup}, \cite{Bitouk:2008}, \cite{kemelmacher2010}, \cite{Yang_siggraph11}), despite the fact that some of these methods are sufficiently advanced to produce natural-looking results. It is also difficult to obtain temporally-coherent results frame-by-frame using existing warping methods (e.g., \cite{Lee:MBA:1997}, \cite{Schaefer:2006}, \cite{StefanoskiWLGHS13}, \cite{Manohar:2008:ICPR}). Face editing in video remains a highly challenging problem, mainly due to complex facial geometry (e.g., fuzzy localization of the eyes on different faces or two successive frames representing the same face) and subtle inaccuracies in facial feature localization (e.g., subtle motion of the eyebrows and facial outline when only the lips are moving between two frames), as illustrated in Fig.~\ref{fig:inconsist}. Even though a temporal-average filter can, to some extent, smooth facial landmarks, it is still difficult to achieve highly consistent localization due to the diversity of facial appearances in different videos, as well as there being sensitivity to filter parameters. Intensive user intervention is therefore usually required for high-performance face editing.

\begin{figure}[!t]
\normalsize
\centering
\subfigure[]{\label{fig:hierarchy}\includegraphics[width=5.3cm]{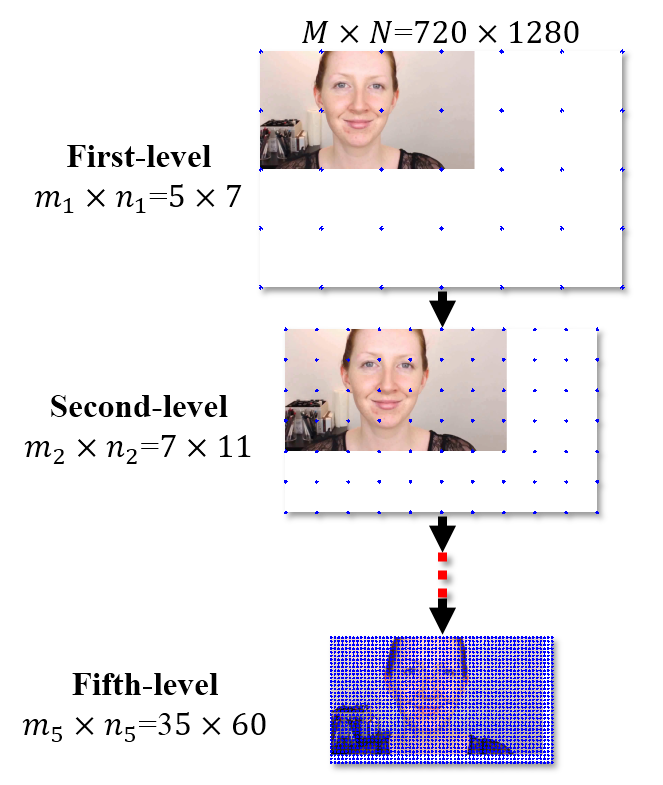}}\hspace{20pt}
\subfigure[]{\label{fig:position}\includegraphics[width=4.8cm]{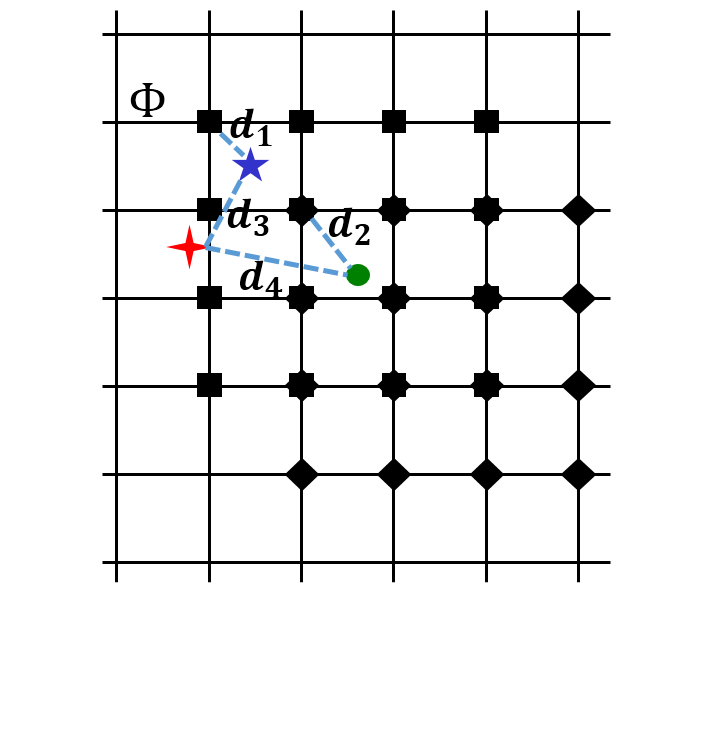}}
\caption{Control lattices and positional relationship between feature and control points. (a) When an image size is {\it M}$\times${\it N}, the size of the {\it h}-level control lattice will be ${\it m_h\times\it n_h}=\lceil{\it M}\times{\it N}/({\rm min}\left\{{\it M},{\it N}\right\}/2^{\it h})\rceil+3$. (b) The 16 diamond control points are selected related to accurate localization of a feature point (marked with the circle), while the 16 squares are chosen by the automatic localization (marked with the star), sharing 9 control points. After face editing, the target position is marked with the cross recording the same feature point.}
\end{figure}

Our approach significantly advances the multilevel B-splines approximation (MBA) \cite{Lee:MBA:1997}, in which control lattices at different levels of a coarse-to-fine hierarchy are repeatedly overlaid on the image plane (Fig.~\ref{fig:hierarchy}). Let $\rm\Phi$ be the control lattice containing many control points on the image plane. Due to temporal incoherence of facial feature localization, the values of control points in the lattice at each level cannot be precisely estimated. In addition, the error gradually increases level by level, and thus MBA performs poorly. Detailed analyses regarding Fig.~\ref{fig:position} are given as follows:

\begin{itemize} 
\item Each feature point is used to find its 16 neighbor control points (e.g., those marked with diamond and square). Inaccurate positions will influence the extraction of related control points in the lattice.
\item The distance (e.g., ${\it d}_{1}$ or ${\it d}_{2}$) between a feature point and its upper-left neighbor is used to compute the cubic B-spline functions, which will assign different values from the coarsest to the finest lattices.
\item Displacement (e.g., ${\it d}_{3}$ or ${\it d}_{4}$) between the source and target points also plays an important role in estimating the values of the control points.
\end{itemize}

\begin{figure}[!ht]
\begin{center}
\includegraphics[width=14cm]{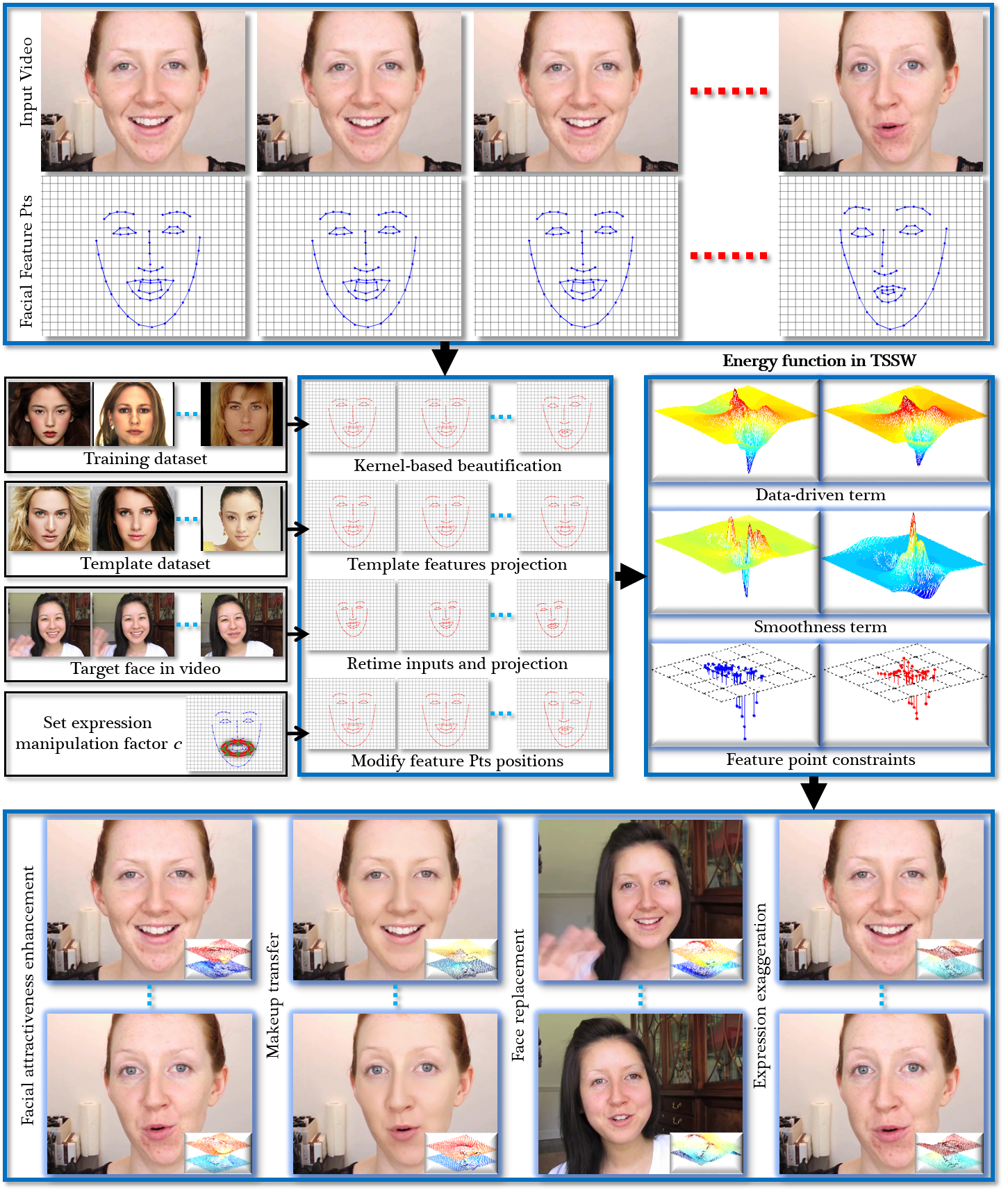}
\caption{A flow chart of video face editing. The resultant image frames are synthesized by the horizontal and vertical warping surfaces obtained by our proposed TSSW method (shown in the bottom-right corner in each example).}
\label{fig:flowchart}
\end{center}
\end{figure}

To overcome the above shortcomings, here we propose a novel temporal-spatial-smooth warping (TSSW) approach to handle temporal incoherence of facial feature localization in videos, which is ignored by MBA or other image-based warping methods. In particular, the goal of TSSW is to achieve temporally-coherent results in a range of video-based face editing applications (Fig.~\ref{fig:flowchart}). The warping task is designed as a total energy minimization problem containing three terms: a data-driven term, a smoothness term, and feature point constraints, which record source and target feature positions of the same face or two different faces and the estimated control lattice (in the horizontal or vertical direction) at its finest level in the previous frame; we only use the control lattice at its finest level since it is known to contain the most available information. The data-driven term minimizes the difference in the values of control points between two successive frames. The smoothness term measures the partial derivatives of the control lattice to be estimated in the current frame, while the feature point constraints enhance one-to-one mapping between two point sets. Once given the estimated control lattices in two directions that record the current frame independently minimized by the total energy function, the corresponding warping surfaces can be computed with cubic B-spline basis functions (see Sec.~\ref{sec:surface}) and used to generate the warped frame. In summary, the proposed method is 

\begin{enumerate}[(1)~]
\item effective to maintain the high temporal coherence in the synthesized videos, and
\item robust to subtle inaccuracies in facial feature localization between two consecutive frames.
\end{enumerate}

The technical description of TSSW and results are available online at {\it our project page}\footnote{\url{https://sites.google.com/site/tsswmethod/}}. The experimental results on four computer vision applications demonstrate that our approach is more convenient and practical than directly applying image-based warping algorithms, and has the potential to be extended for other computer graphics applications.

The remainder of this paper is organized as follows. Section 2 reviews the related work. In Section 3, we introduce the main steps in general video face editing. Section 4 presents the proposed TSSW approach. Section 5 shows the experimental results of four applications and compares our proposed approaches with the state-of-art methods. Section 6 discusses the conclusions, limitations, and future work.

\section{Related Work}
The proposed TSSW can be applied to and improve the performance of various video face editing tasks, which is related to previous research in the following fields.

\begin{itemize} 
\item {\it Image warping} is used to retain 2D geometric transformations between feature point pairs.
\begin{enumerate}[(i)~]
\item Traditional image warping methods include radial basis functions (RBF) \cite{Arad:RBF:1994} and thin plate splines (TPS) \cite{Donato:TPS:2002}. Multilevel free-form deformation (MFFD) \cite{Lee:MFFD:1996} and its improved version by applying B-spline refinement to the control lattices in terms of a coarse-to-fine hierarchy \cite{Lee:MBA:1997}, are presented for image metamorphosis.
\item Moving least squares (MLS) \cite{Schaefer:2006} is designed to deform images using rigid transformation, which tends to make the image deformation as-rigid-as-possible. To focus on nonrigid image deformation, moving regularized least squares (MRLS) technique \cite{Ma:MRLS:2013} interpolates a nonlinear function derived from the scattered data.
\end{enumerate}
However, the image-based warping methods often ignore the temporal coherence for video-based applications.

\item {\it The 3D-based approaches} try to create 3D facial models from 2D images, in which it is challenging to detect all the facial components, such as eyes and mouth. 
\begin{enumerate}[(i)~]
\item The 3D-aware appearance optimization technique is applied to face morphing \cite{Yang:GI:2012} and face component transfer \cite{Yang_siggraph11}. In addition, the 3D morphable face models can be used to enhance the symmetry and proportion of face geometry \cite{Liao:2012}, and suggest the best-fit makeup for an input human face \cite{Scherbaum:2011:Makeup}. 
\item \cite{Dale:2011} and \cite{Yang:2012} extend Vlasic's 3D tensor approach \cite{Vlasic:2005} to edit the facial performances of one or two identities in videos. A FaceWarehouse database \cite{Chen:2014} consisting of 3D facial expressions is constructed for various computer vision applications.
\end{enumerate}
However, the 3D data is usually difficult to obtain and requires considerable user interactions on key-frames to produce accurate identity parameters. In addition, the computational cost is high for using 3D models.

\item {\it Shape registration} aims to construct optimal transformation for shapes of interest. 
\begin{enumerate}[(i)~]
\item An efficient registration method based on a cubic deformation model \cite{Taron:TPAMI:2009} is designed to recover a one-to-one correspondence between source and target shapes. To solve the explicit shape matching problem, a dual decomposition approach \cite{Torresani:TPAMI:2013} is proposed to establish the correspondences between sparse image feature points.
\item \cite{Huang:2006} and \cite{Abd:TPAMI:2013} estimate the global and local transformation parameters using implicit distance function and energy optimization.
\end{enumerate}
Even though the smoothness term in our total energy function is similar to the field of shape registration, the image-based registration methods cannot properly preserve temporal correspondences between the lattices and edit foreground objects (e.g., face region), which is thus not practical for video face editing. In addition, the optimal transformation parameters recovered by the registration methods are computationally expensive and unstable over a sequence of video frames.

\item {\it Video warping} is used to render each video frame based on the deformed grid mesh recording the optimized feature positions when applied to video retargeting and video stabilization. 
\begin{enumerate}[(i)~]
\item \cite{Wang:TOG:2010} and \cite{Lin:TVCG:2013} exploit several spatial deformation (e.g., nonuniform global mesh warping) and temporal coherence constraints to preserve visually salient content (e.g., foreground objects) for video retargeting. 
\item Video stabilization techniques have been developed to smooth shaky camera motions, such as structure from motion (SFM) model proposed by \cite{Liu:CPW:2009} and spatial-temporal optimization method \cite{Wang:TVCG.2013}. 
\end{enumerate}
However, neither preserving the foreground objects nor stabilizing shaky camera motions in warping is effective for editing facial components in the foreground. Indeed, the cameras used in our experiments are fixed and the proposed video face editing not only edits facial components in each frame, but also preserves the temporal coherence.

\item {\it Image face editing} has gained extensive attention in recent years.
\begin{enumerate}[(i)~]
\item A face attractiveness enhancement engine \cite{Leyvand:2008} is presented to modify the distances between facial feature positions by exploring those of a set of training faces. A framework proposed by \cite{Guo:2009} is designed to mock physical makeup by creating the makeup upon a face through a template image. 
\item A system \cite{Bitouk:2008} is introduced to swap faces by finding candidates similar to the appearance and pose of the input face from a large-scale face dataset. \cite{Yang_siggraph11} replaces two different face expressions between two photographs of the same person using the optical flow derived from 3D morphable models.
\end{enumerate}
However, these schemes cannot achieve temporally coherent results for video face editing applications.
\end{itemize}

\section{Face Editing in Video}
Fig.~\ref{fig:flowchart} shows a flow chart of video face editing containing four main steps, which exploits the proposed TSSW method. First, given an input video, 2D facial feature localization (Sec.~\ref{sec:feature}) should be performed on the original image frames because this information guides subsequent editing of facial components, and therefore feature localization with relative accuracy is necessary. Second, facial components are then edited (Sec.~\ref{sec:editing}) according to the desired application (e.g., facial attractiveness enhancement, makeup transfer, face replacement, and expression manipulation) facilitated by some necessary data. The original and modified feature points are then sent to TSSW. Third, according to the feature point pairs and the horizontal and vertical control lattices in the previous frame, the two corresponding control lattices in the current frame are estimated by minimizing an energy function (Sec.~\ref{sec:energy}). The corresponding control lattices together with the cubic B-spline basis functions generate the warping surfaces (Secs.~\ref{sec:warp} and~\ref{sec:surface}). Finally, the warped frames obtained by TSSW deform the facial components followed by the post-processing (e.g., Poisson blending in face replacement is used to patch the face boundary in the synthesized result).

\subsection{Facial Feature Localization}
Since our warping algorithm is based on the landmarks found in 2D face geometry, we improve Supervised Descent Method (SDM) proposed by \cite{Xiong:2013:ICCV}, named as ISDM, to automatically detect and track 66 facial feature points for each frame (while 49 feature points used in \cite{Xiong:2013:ICCV}). The feature points, often located in the detected face region, infer five facial components: (left and right) eyebrows, (left and right) eyes, nose, (upper and lower) lips, and facial outline (Fig. 1(b)) \cite{Luo:2013:ICCV}. Due to the diversity of faces possible in various applications, the feature points are roughly located by an automatic tracking technique (e.g., \cite{Candamo:2009:TAES}, \cite{Sun:2013:ICCV}, \cite{Song:2013:ACMTIST:OSM}), and then refined, thus this takes several minutes for each frame; however, refinement and use of a smoothing filter only offer marginal improvement, and temporally coherent feature localization is difficult to perform on the whole video. In addition, very subtle differences in facial landmarks between two consecutive frames, which are often perceptually insensitive to the Human Visual System (HVS), affect the synthesized results in face editing applications.
\newline \indent TSSW concentrates efforts by constructing control lattices in each frame for warp generation, regardless of subtle differences in facial feature localization. In other words, users are not required to refine the position of the feature points to achieve high accuracy; instead ISDM is directly applied. However, since facial feature tracking relies on optical flow, initialization of facial features is critical because errors will be propagated and then accumulated frame by frame. We therefore assume that the initialization of tracking is good for each input video.
\label{sec:feature}

\subsection{Facial Component Editing}
In this paper we demonstrate four scenarios: facial attractiveness enhancement, makeup transfer, face replacement, and expression manipulation. Since our framework is based on 66 facial features, the goal of editing facial components is to obtain modified facial landmarks. It is well known that different applications should use different editing engines: 1) for facial attractiveness enhancement, a kernel-based facial attractiveness enhancement method is proposed to construct a set of new facial landmarks for the input frames with the help of a training dataset; 2) for transferring makeup, a similar template face to be warped (with similar skin color to that of the input face) is selected from a template dataset. The feature points of the selected template are then projected on the original frame as the new positions; 3) for face replacement, the input video sequence is required to retime on demand to match the expressions and pose information in the target video. Furthermore, the feature points of each retimed frame are mapped onto the corresponding target frame by projection transformation (e.g., affine transformation); and 4) for manipulating facial expressions in the whole video, given a manipulation factor, the feature point positions of the (upper and lower) lips and chin are modified and then projected onto the corresponding retimed frame as the new positions.
\newline \indent Overall, the four applications use four different facial component editing engines in order to obtain the feature point pairs, which play a critical role in the subsequent warping process.
\label{sec:editing}

\subsection{Warp Generation}
Warp generation is performed according to feature point pairs in order to further map a set of original facial feature points to the modified ones in the synthesized frame. There are many different types of warp generation, including affine, similarity, rigid transformations, and other sophisticated transformations (e.g., \cite{StefanoskiWLGHS13} and \cite{Siddiqui:2009}). The classical image-based warping methods are MBA and MLS, but indeed these two methods cannot produce realistic and temporally coherent results, especially when there is a complicated background in video. In addition, MBA has several limitations (discussed in Sec.~\ref{sec:introd}) when there is temporal incoherence of facial feature localization.

For a good initialization, we first compute two control lattices in the horizontal and vertical directions (since each feature point has displacement in these two directions) at the first frame using MBA. The control lattices in the subsequent frames are then estimated by the proposed energy function (see Sec.~\ref{sec:tssw}), using temporal information from the control lattices in each preceding frame. The corresponding warping surfaces are generated and further guided to create new facial components from the warped version of each frame according to one of various different face-editing applications.
\label{sec:warp}

\section{Temporal-Spatial-Smooth Warping}
\label{sec:tssw}
The input of TSSW is a video containing $T$ frames of a human face. Applying ISDM, the 2D facial feature points in a frame are denoted as ${\it Q=\left \{(x_k,y_k) \right \}_{k={\rm 1}}^K}$, which consists of the x- and y-coordinates of its {\it K} landmarks. The modified feature points are defined as {\it P}, which also possesses two coordinates representing its {\it K} facial positions. For each frame ${\it t}$ $(1\leq{\it t}\leq T)$, the feature point pairs can be represented as (${\it Q_t,P_t}$). TSSW precisely estimates the control lattices from the second to the last frame. The estimated control lattices are then used to generate the corresponding warping surfaces.

\subsection{Description of Warping Surfaces}
\label{sec:surface}
Assuming a finest control lattice at the {\it H}-level, the size of the finest control lattice is ${\it m_H\times\it n_H}$ (see Fig.~\ref{fig:hierarchy}) with a scaling factor ${\it s_H}$. Let the size of the image plane be ${\it M\times N}$. The image points are then scaled as ${\it x_{H,k}=s_{H}\cdot(x_k-{\rm1})+{\rm 1}}$ and ${\it y_{H,k}=s_{H}\cdot(y_k-{\rm1})+{\rm 1}}$ for $1\leq{\it x_k}\leq{\it M}$, $1\leq{\it y_k}\leq{\it N}$. Once given each control lattice, the {\it k}-th warping surface value can be calculated by

\vspace{-6pt}
\begin{align}
{\it f_k}({\rm \Psi}_{\it t}^{\it l}) & =\sum_{{\it i}=0}^3\sum_{{\it j}=0}^3{\it B_i}({\it u_k}){\it B_j}({\it v_k}){\rm \Psi}_{\it t}^{\it l}({\it i+i_k},{\it j+j_k}),
\label{eq:sur}
\end{align}
\vspace{-6pt}

\noindent where ${\it i_k}=\lfloor{\it x_{H,k}}\rfloor$, ${\it j_k}=\lfloor{\it y_{H,k}}\rfloor$, ${\it u_k}={\it x_{H,k}}-{\it i}_0$, and ${\it v_k}={\it y_{H,k}}-{\it j}_0$. $\lfloor\cdot\rfloor$ is the rounded down operation. The horizontal warping surface value is computed for ${\it l}=1$, while the vertical warping surface value is computed for ${\it l}=2$. In addition, the four-order cubic B-spline basis functions $\left\{{\it B_i}(\cdot)\right\}_{{\it i}=0}^3$ are defined, as illustrated in \cite{Lee:MBA:1997}:

\vspace{-6pt}
\begin{align}
{\it B_i}({\it u}) & = {\it a_i}[{\it u}^3\;{\it u}^2\;{\it u}^1\;{\it u}^0]^{\mathscr{T}},\;\;\;{\it i}=0,1,2,3,
\label{eq:bsp}
\end{align}
\vspace{-6pt}

\noindent where $0\leq{\it u}<1$. $\left\{{\it a_i}\right\}_{{\it i}=0}^3$ are the basis vectors: ${\it a}_0=[-1\;3\;-3\;1]/6$, ${\it a}_1=[3\;-6\;0\;4]/6$, ${\it a}_2=[-3\;3\;3\;1]/6$, and ${\it a}_3=[1\;0\;0\;0]/6$. The symbol of $\mathscr{T}$ represents the transpose operation. The cubic B-spline basis functions are considered to weigh the contribution of its 16 neighbor control points based on the distance to its upper-left control point (e.g., ${\it d}_1$ and ${\it d}_2$, as shown in Fig.~\ref{fig:position}).
\newline \indent Using matrix notation, we rewrite Eq.~(\ref{eq:sur}) as

\vspace{-6pt}
\begin{align}
{\it f_k}({\rm \Psi}_{\it t}^{\it l}) & ={\it W_k}^{\mathscr{T}}{\rm \Psi}_t^l,\;\;\;{\it for\;each}\;({\it x_k, y_k}),
\label{eq:matrix}
\end{align}
\vspace{-6pt}

\noindent where ${\it W_k}$ is a weighted matrix with respect to the corresponding image point (${\it x_k, y_k}$), which is defined as follows:

\vspace{-6pt}
\begin{align}
 {\it W_k(i+i_k,j+j_k)}=\left\{\begin{matrix}
{\it B_i}({\it u_k}){\it B_j}({\it v_k}), & 0\leq{\it i,j}\leq3,\\
0, & {\it otherwise.}
\end{matrix}\right.
\label{eq:weight}
\end{align}
\vspace{-6pt}

The horizontal and vertical warping surfaces share the same weighted matrix. Using these formulations, the problem of deriving the warping surfaces is reduced to solving only each control lattice that relates to the feature point pairs. Since the goal is to estimate control lattices with temporal information, we therefore directly use MBA to obtain the control lattices (at the sixth level in the experimental setting) in the first frame for optimal initialization. The subsequent control lattices are then estimated using TSSW. We can therefore obtain the horizontal and vertical warping surfaces for each frame using the above equations.

\subsection{Energy Function}
\label{sec:energy}

From the second to the last frame, i.e., ${\it t}>1$, TSSW is designed as a problem of energy function minimization to estimate the current control lattice ${\rm \Psi}_{\it t}$ with the temporal information of the previous control lattice ${\rm \Psi}_{{\it t}-1}$. The total energy function is formulated as the weighted sum of three terms:

\vspace{-6pt}
\begin{align}
{\it E({\rm \Psi}_t^l)=E_d({\rm \Psi}_t^l)+\alpha E_s({\rm \Psi}_t^l)+\beta E_f({\rm \Psi}_t^l)},\;\;{\it t}>1,
\label{eq:energy}
\end{align}
\vspace{-6pt}

\noindent where ${\it \alpha}>0$ and ${\it \beta}>0$ are two regularization parameters that balance the tradeoff between data-driven term ${\it E_d({\rm \Psi}_t^l)}$, smoothness term ${\it E_s({\rm \Psi}_t^l)}$, and feature point constraints ${\it E_f({\rm \Psi}_t^l)}$. For frame ${\it t}$, the control lattices in two directions (i.e., horizontal and vertical directions) can be estimated using the same energy function (${\it l}=1$ and ${\it l}=2$), which can be minimized using the conjugate gradient technique. The descriptions of the three terms are shown as follows.

\subsubsection{Data-driven term}Compared with the temporal smoothness regularization term \cite{Liu:2013:ACMTIST:STC} which is based on the collection of all the time dependent model parameters, we only conduct the function minimization on two consecutive video frames. The data-driven term penalizes the difference between the current control lattice ${\it {\rm \Psi}_t^l}$ and the previous control lattice ${\it {\rm \Psi}_{t-{\rm 1}}^l}$ using the sum-of-squared-differences (SSD) criterion. Given the control lattice in the previous frame, ${\it {\rm \Psi}_{t-{\rm 1}}^l}({\it t}>1)$, we define the data-driven term as

\begin{align}
{\it E_d({\rm \Psi}_t^l)=\sum_{i={\rm1}}^{m_H}\sum_{j={\rm1}}^{n_H}\left({\rm \Psi}_t^l(i,j)-{\rm \Psi}_{t-{\rm 1}}^l(i,j)\right)^{\rm2}.}
\label{eq:data}
\end{align}

The modified feature points are also in the face region. According to the feature point pairs, the data-driven term is designed to change the control lattice in the face region but not in the background. This term guarantees that the values of the control points have as small differences as possible between two consecutive frames in order to achieve temporally coherent results in the final optimization.

\subsubsection{Smoothness term} The smoothness term preserves the regularity of the current control lattice to be estimated using its gradients. Specifically, for each ${\rm \Psi}_{\it t}$ on the same plane ${\rm \Phi}$, an efficient smoothness term is defined as

\begin{align}
{\it E_s({\rm \Psi}_t^l)}=\sum_{i={\rm1}}^{m_H}\sum_{j={\rm1}}^{n_H}\left(\nabla_x^{\rm2}(i,j)+\nabla_y^{\rm2}(i,j)\right),
\label{eq:smooth}
\end{align}

\noindent where $\nabla_{\it x}=\partial\Psi_{\it t}^{\it l}/\partial {\it x}$ and $\nabla_{\it y}=\partial\Psi_{\it t}^{\it l}/\partial {\it y}$ denote the first-order partial derivatives of ${\it {\rm \Psi}_t^l}$.

Such a smoothness term can be further based on an error norm with some known limitations. However, \cite{Huang:2006} suggest that an implicit smoothness constraint imposed by cubic B-spline basis functions can guarantee first derivative continuity on the control points, while the second derivative has continuity elsewhere. Therefore, we directly integrate the smoothness term (Eq.~(\ref{eq:smooth})) into the energy function to recover the current control lattice ${\it {\rm \Psi}_t^l}$, as the computation of the control lattice is based on cubic B-spline basis functions. Moreover, the control lattice to be estimated should itself have a smooth texture in the image field.

\subsubsection{Feature point constraints} The data-driven and smoothness terms in the energy function are suboptimal if the estimation error is large. To address this problem, we impose feature point constraints. We then model the feature point constraints as an SSD measure between each warping surface on the modified facial feature points ${\it P_t}$ and the 2D geometric displacements derived from the feature point pairs (${\it Q_t,P_t}$). Therefore, the energy function incorporates the following feature point constraints:

\begin{align}
{\it E_f({\rm \Psi}_t^l)=\sum_{(x_k,y_k)\in P_t}\left(f_k({\rm \Psi}_t^l)-z_{t,k}^l\right)^{\rm2}.}
\label{eq:pairs}
\end{align}

Denote ${\it Z_t=\{(z_{t,k}^{\rm1},z_{t,k}^{\rm2})\}_{k={\rm1}}^K=Q_t-P_t}$. In Eq.~(\ref{eq:pairs}), regarding the {\it k}th facial feature point, ${\it z}_{\it t,k}^{\rm1}$ and ${\it z}_{\it t,k}^{\rm2}$ represent the displacements in the horizontal and vertical directions, respectively. In addition, ${\it f_k(\cdot)}$ is the value of the warping surface (Eq.~\ref{eq:matrix}) recording the {\it k}th modified feature point in ${\it P_t}$.
\newline \indent Feature point constraints are critical and greatly improve the accuracy and efficiency in estimating each control lattice. Hence, the regularization parameter $\beta$ is set to a larger value than $\alpha$ in Eq.~(\ref{eq:energy}).

\subsection{Optimization}

The total energy function (in Eq.~(\ref{eq:energy})) is a quadratic function of the control lattice ${\it {\rm \Psi}_t^l}$. Combining Eqs.~(\ref{eq:data})-(\ref{eq:pairs}) into Eq.~(\ref{eq:energy}) attains a local minimum when ${\it {\rm \Psi}_t^l}$ satisfies the following Euler-Lagrange equation:

\begin{align}
{\it \frac{\partial E}{\partial {\rm \Psi}_t^l}-\frac{\partial}{\partial x}\frac{\partial E}{\partial {\rm \nabla}_x}-\frac{\partial}{\partial y}\frac{\partial E}{\partial {\rm \nabla}_y}=}0.
\label{eq:total}
\end{align}

\noindent in which:

\vspace{-6pt}
\begin{align}
{\it \frac{\partial E}{\partial {\rm \Psi}_t^l}} = 2{\it\left({\rm \Psi}_t^l-{\rm \Psi}_{t-{\rm1}}^l\right)}+2{\it\beta\sum_{k={\rm1}}^KW_k\left(W_k^{\mathscr{T}}{\rm \Psi}_t^l-z_k^l\right)},
\end{align}
\vspace{-15pt}

\begin{align}
{\it\frac{\partial}{\partial x}\frac{\partial E}{\partial {\rm \nabla}_x}}=2{\it\alpha\frac{\partial^{\rm2}{\rm \Psi}_t^l}{\partial x^{\rm2}}},\quad and \quad
{\it\frac{\partial}{\partial y}\frac{\partial E}{\partial {\rm \nabla}_y}}=2{\it\alpha\frac{\partial^{\rm2}{\rm \Psi}_t^l}{\partial y^{\rm2}}}.
\end{align}

\begin{algorithm}[!t]
\SetAlgoNoLine
\KwIn{The original video frames $\{X_{in}^t\}_{t=1}^T$ and the corresponding facial feature point pairs $\{\left(Q_t,P_t\right)\}_{t=1}^T$, and two regularization parameters $\alpha$ and $\beta$.}
\KwOut{The warped video frames $\{X_{warp}^t\}_{t=1}^T$.}
{\bf Initialization:} $t$ = 1; Obtain $X_{warp}^1$ by MBA method and the control lattices $\{\Psi_1^l\}_{l=1,2}$\;
$t=t+1$\;
Compute the Laplacian matrix $L$ with size of $m_Hn_H\times m_Hn_H$\;
\For{$t>1$}{
Calculate $\sum_{k=1}^KW_kW_k^{\mathscr{T}}$ regarding $P_t$ using Eq.~(\ref{eq:weight}) and Eq.~(\ref{eq:bsp})\;
\For{$l=1,2$}{
Calculate $\sum_{k=1}^KW_kz_{t,k}^l$ with the geometric displacements derived from $\left(Q_t,P_t\right)$\;
\Repeat{reach the tolerance or the maximum number of iterations}{
Calculate Eq.~(\ref{eq:linear}) with the previous control lattice $\Psi_{t-1}^l$ using conjugate gradient method\;
}
Obtain the current control lattice $\Psi_t^l$\;
}
Calculate the $M\times N$ warping surfaces for each pixel position with $\{\Psi_t^l\}_{l=1,2}$ using Eq.~(\ref{eq:sur})\;
Obtain the corresponding warped frame $X_{warp}^t$ by mapping $X_{in}^t$ with the warping surfaces $\{f(\Psi_t^l)\}_{l=1,2}$ using bicubic interpolation method\;
$t=t+1$.
}
\caption{Temporal-Spatial-Smooth Warping (TSSW) Method}
\label{alg:tssw}
\end{algorithm}

Combining the above equations, Eq.~(\ref{eq:total}) can be rewritted as

\vspace{-10pt}
\begin{align}
{\it \left({\rm \Psi}_t^l-{\rm \Psi}_{t-{\rm1}}^l\right)+\beta\sum_{k={\rm1}}^K\left(A_k{\rm \Psi}_t^l-W_kz_{t,k}^l\right)-\alpha L{\rm \Psi}_t^l}=0,
\end{align}

\noindent where ${\it A_k=W_kW_k^{\mathscr{T}}}$ and ${\it L=D_x^{\mathscr{T}}D_x+D_y^{\mathscr{T}}D_y}$. ${\it L}$ is the homogeneous Laplacian matrix. ${\it D_x}$ and ${\it D_y}$ represent the forward difference operators. ${\it A_k} (1\leq{\it k}\leq{\it K})$ and ${\it L}$ are both symmetric.

After reorganizing and simplifying, the resulting linear system for ${\it {\rm \Psi}_t^l}$ is given by

\begin{align}
\left(I+\beta\sum_{k=1}^KA_k-\alpha L\right)\Psi_t^l=\left(\Psi_{t-1}^l+\beta\sum_{k=1}^KW_kz_{t,k}^l\right),
\label{eq:linear}
\end{align}
where ${\it I}$ is an identity matrix with size of $m_Hn_H\times m_Hn_H$.

Through the cubic B-spline basis functions (Eq.~(\ref{eq:bsp})), the estimation of ${\it {\rm \Psi}_t^l}$ is only related to the previous control lattice ${\it {\rm \Psi}_{t-{\rm1}}^l}$ and the feature point pairs obtained from the facial components editing process. The linear system (Eq.~(\ref{eq:linear})) can then be optimized via a sequence of conjugate gradient iterations. Note that between consecutive iterations, the control lattice can be gradually updated with highly temporally coherent information. The TSSW process is summarized in Algorithm 1.

Once given the control lattices $\{\Psi_t^l\}_{l=1,2}$ in the current frame, the corresponding warping surfaces are computed using Eq.~(\ref{eq:matrix}), which lead to more temporally coherent and spatially smoother, and therefore more realistic results.

\section{Applications and Results}

We implement and test the proposed approaches on an Intel Core 2 Duo 3.0 GHz CPU and 4 GB memory in the Matlab environment. To show TSSW is crucial for the high performance video face editing, we conduct a number of validations for the four applications: facial attractiveness enhancement, makeup transfer, face replacement, and expression manipulation. Table~\ref{lab:video} shows the information of videos from {\it YouTube website}\footnote{https://www.youtube.com/} used in our work and the corresponding timing statistics obtained by our methods. For all experiments, the two regularization parameters in Eq.~(\ref{eq:linear}) are set to ${\it\alpha}=0.8$ and ${\it\beta}=1$. Using conjugate gradient technique, the optimization step is iterated about 30 times. Since the size of face region in different videos may vary widely for the same application, the runtime for two different videos is significantly different. The details of the implementation in the above four applications are illustrated below, and the overall results are available online\footnote{\url{http://youtu.be/LQCLeQcBS74}}.

\begin{table}
\footnotesize
\newcommand{\tabincell}[2]{\begin{tabular}{@{}#1@{}}#2\end{tabular}}
\begin{center}
\caption{Video Information and Runtime (seconds) obtained by the proposed method}{
\subtable[Attractiveness and Makeup]{
\begin{tabular}{|c|c|ccc|} \hline
\multicolumn{2}{|c|}{Name} & Resolution &  NoF  & TPF \\ \hline
\multirow{3}{*}{\tabincell{c}{Attract- \\ iveness}} &
Video 1 & $480 \times 856$ & 301 & 6.269	\\
\cline{2-5} & Video 2 & $360 \times 640$ & 3000 & 4.412	\\
\cline{2-5} & Video 3 & $720 \times 1280 $ & 144 & 6.097	\\ \hline
\multirow{3}{*}{Makeup} & 
Video 4 & $720 \times 1280 $ & 115 & 19.211	\\
\cline{2-5} & Video 5 & $720 \times 1280 $ & 250 & 22.934	\\
\cline{2-5} & Video 6 & $720 \times 1280 $ & 1625 & 15.755	\\ \hline
\end{tabular} }
\subtable[Replacement and Manipulation]{
\begin{tabular}{|c|c|ccc|} \hline
\multicolumn{2}{|c|}{Name} & Resolution &  NoF  & TPF\\ \hline
\multirow{3}{*}{\tabincell{c}{Repla- \\ cement}} & 
Video 7 & $704 \times 1243 $ & 699 & 23.063	\\
\cline{2-5} & Video 8 & $720 \times 1280 $ & 88 & 17.193	\\
\cline{2-5} & Video 9 & $688 \times 1280 $ & 86 & 57.632	\\
\hline
\multirow{3}{*}{\tabincell{c}{Manip- \\ ulation}} & 
Video 10 & $720 \times 1280$ & 2000 & 7.017	\\
\cline{2-5} & Video 11 & $720 \times 1280$ & 150 & 6.001	\\
\cline{2-5} & Video 12 & $720 \times 1280$ & 699 & 6.882	\\
\hline
\end{tabular} } }
\vskip-5pt
\begin{itemize}
\item \quad NoF represents the total number of video frames.
\item \quad TPF stands for the average runtime per frame.
\item \quad The timing of feature detection and tracking is not included in this table.
\item \quad Regarding replacement, this table only shows the information of the target video.
\end{itemize} 
\label{lab:video}
\end{center}
\end{table}

\begin{figure}[!t]
\begin{center}
\includegraphics[width=10.8cm]{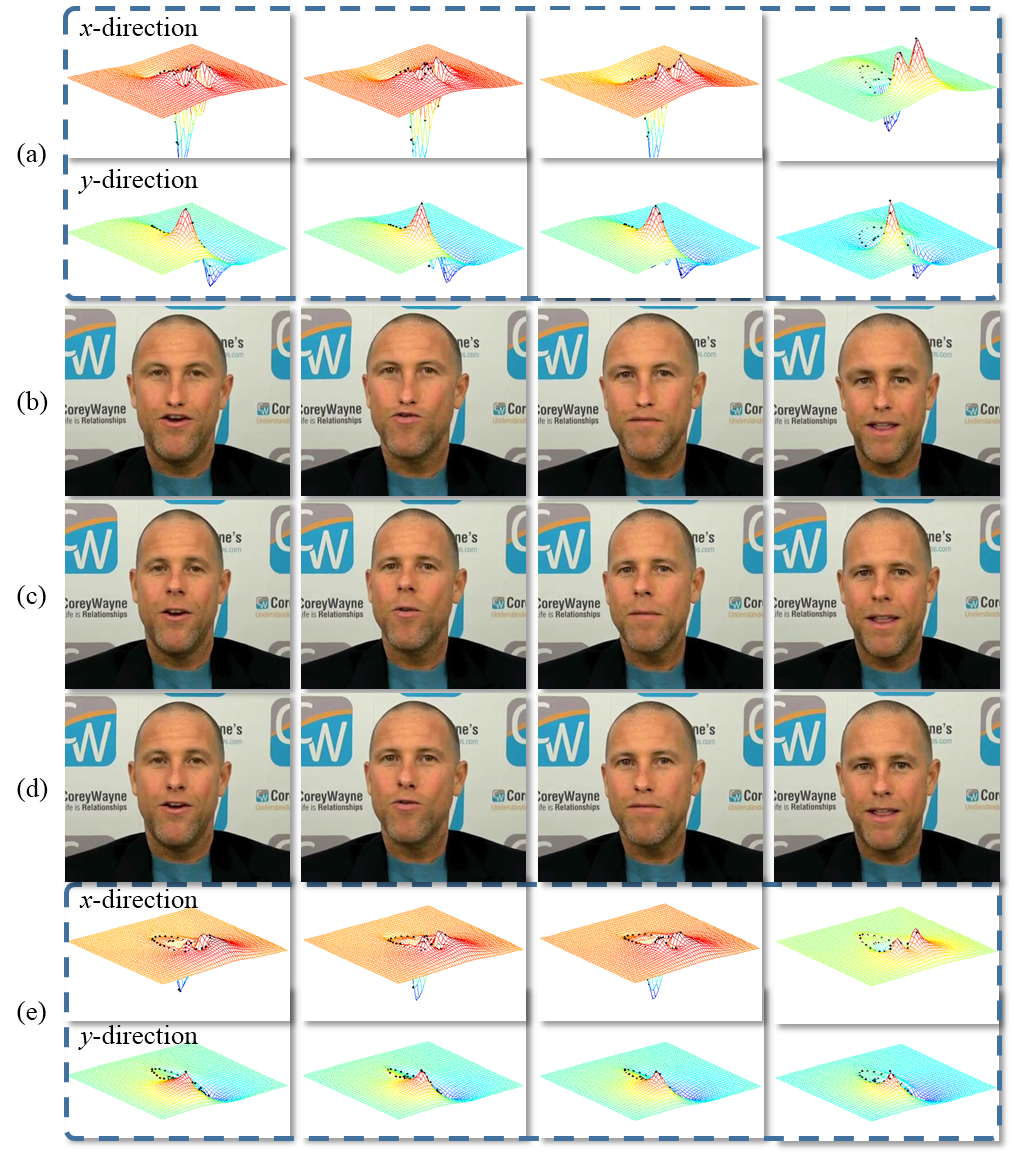}
\caption{ Facial attractiveness enhancement and comparison of warping surfaces ({\bf Video 2}). (c) Original frames. (a) and (b): Warping surfaces and synthesized frames obtained using MBA. (d) and (e): Our synthesized frames and warping surfaces.}
\label{fig:surface}
\end{center}
\end{figure}

\begin{figure}[!t]
\centering
\subfigure[]{\includegraphics[width=2.7cm]{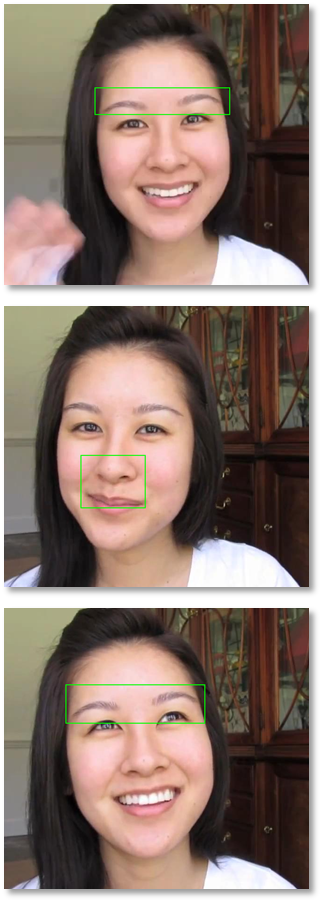}}\hspace{-3pt}
\subfigure[]{\includegraphics[width=2.7cm]{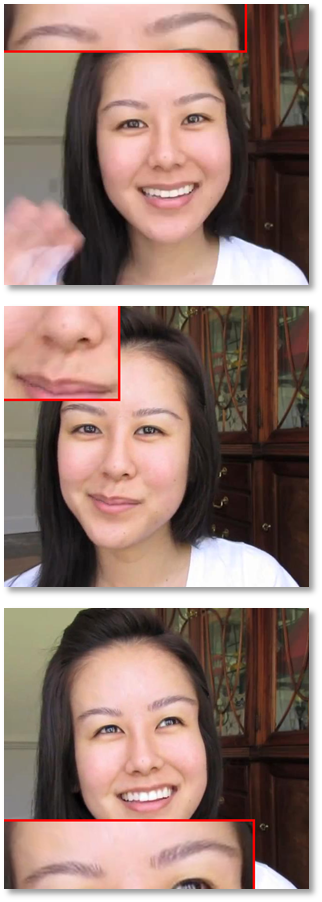}}\hspace{-3pt}
\subfigure[]{\includegraphics[width=2.7cm]{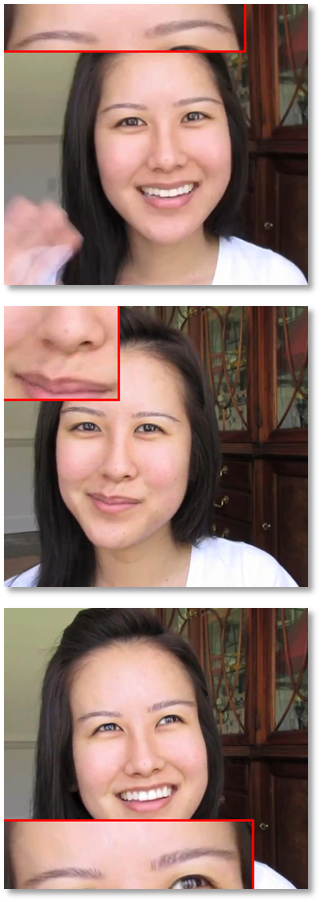}}\hspace{-3pt}
\subfigure[]{\includegraphics[width=2.7cm]{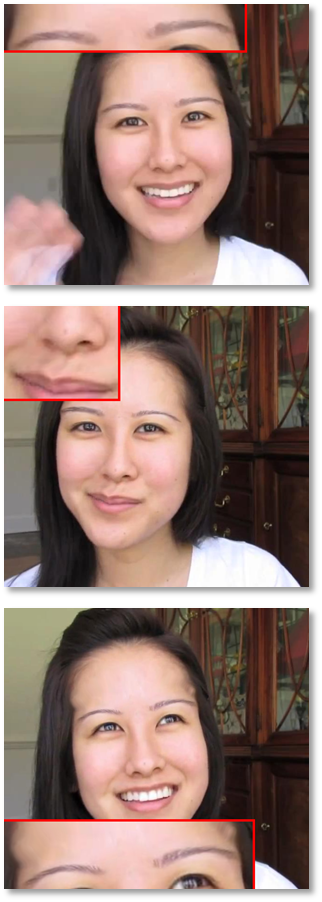}}\hspace{-3pt}
\subfigure[]{\includegraphics[width=2.7cm]{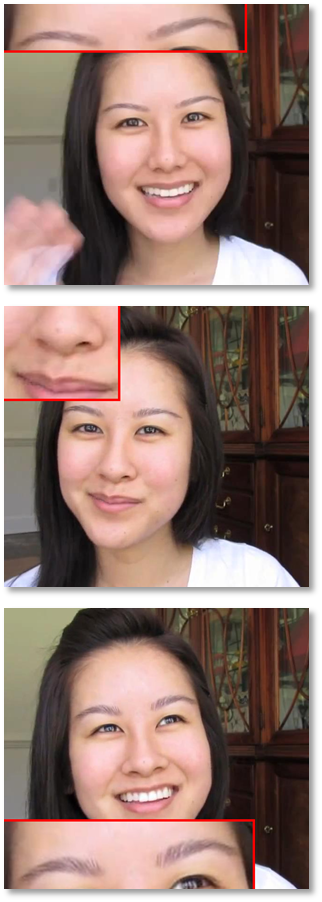}}
\caption{Comparison of facial attractiveness enhancement results ({\bf Video 3}). (a) Original frames. (b) \cite{Leyvand:2008}. (c) MBA \cite{Lee:MBA:1997} + kernel. (d) MLS \cite{Schaefer:2006} + kernel. (e) Ours.}
\label{fig:comp:attract}
\end{figure}

\subsection{Facial Attractiveness Enhancement}
A training dataset of neutral faces (101 female faces and 94 male faces) are used in our construction. In this experiment, we use the 66 facial feature points obtained by ISDM, while \cite{Leyvand:2008} requires a total of 84 feature points. We utilize global symmetrization and overall proportion optimization \cite{Liao:2012} on key facial features to calculate the beauty score of a face. A higher beauty score corresponds to a more attractive face in the training dataset. The feature points of a training face construct a 174-dimensional distance vector using the Delaunay triangulation. Due to the differences in face geometry between females and males, we construct two training databases, i.e., $\{(\mathbb{V}_{s,l}^i,b_{i,l})\}_{i=1}^{n_{s,l}}$, in which $\mathbb{V}_{s,l}^i$ is the distance vector, $b_{i,l}$ is the corresponding beauty score, $l=1$ for females, $l=2$ for males, and $n_{s,l}$ is the total number of faces in the $l$-th training subset.

In this paper, we propose a kernel-based scheme for facial attractiveness enhancement. Given a video, we first confirm the gender of the input face i.e., $l$. The original feature points captured by ISDM are denoted as $\{Q_i\}_{i=1}^T$. Similar to the construction of training database, we calculate the corresponding distance vectors, i.e., $\{\mathbb{V}_i\}_{i=1}^T$. Then, we find the frame with neutral expression and the corresponding distance vector (denoted as ${\it v_{neu}}$) by measuring eyes and mouth opening distances. Based on the corresponding training subset, the similarity weight ${\it w_{ij}}$ for each distance vector $\mathbb{V}_i$, as

\begin{align}
w_{ij} = b_{j,l}\cdot {\rm exp}\left(-\frac{\left\|\mathbb{V}_i-\mathbb{V}_{s,l}^j\right\|_2^2}{\sigma^2}\right),
\label{eq:kernel}
\end{align}
where $\sigma$ is the kernel parameter in the weighting computation ($\sigma=5$ in this experiment). From Eq.~(\ref{eq:kernel}), the best beautiful face may not be extracted for all the input faces. Large weight relates to high beauty score, as well as small difference between distance vectors. The new distance vector $\mathbb{V}_i^{\rq}$ with higher attractiveness rating can be computed as

\begin{align}
\mathbb{V}_i^{\rq} = \frac{\sum\limits_{j\in \Gamma\left(v_{neu}\right)}w_{ij}\mathbb{V}_{s,l}^j}{\sum\limits_{j\in \Gamma\left(v_{neu}\right)}w_{ij}},
\end{align}
where $\Gamma\left({\it v_{neu}}\right)$ represents the similar neighbors that are relatively close to the neutral distance vector ${\it v_{neu}}$. The similar neighbors are used for all the input video frames. In this experiment, the number of neighbors is set to 5.

A set of new facial feature points $\{P_i\}_{i=1}^T$ is inferred from the modified distance vectors using Levenberg-Marquardt (LM) algorithm \cite{marquardt:1963}. The warping surfaces are generated by minimizing the proposed energy function. Fig.~\ref{fig:surface} shows the comparison of horizontal and vertical warping surfaces, which demonstrates that TSSW only modifies the warping surface values in the face region and preserves temporal coherence between video frames. Fig.~\ref{fig:comp:attract} and the online video demonstration\footnote{\url{http://youtu.be/8ZYUXlNpeOg}} suggest TSSW is superior to \cite{Leyvand:2008} and the combination methods (e.g., MBA method with our kernel-based attractiveness enhancement engine) by showing the synthesized results of facial attractiveness enhancement.

\begin{figure}[!t]
\normalsize
\centering
\subfigure[]{\includegraphics[width=3cm]{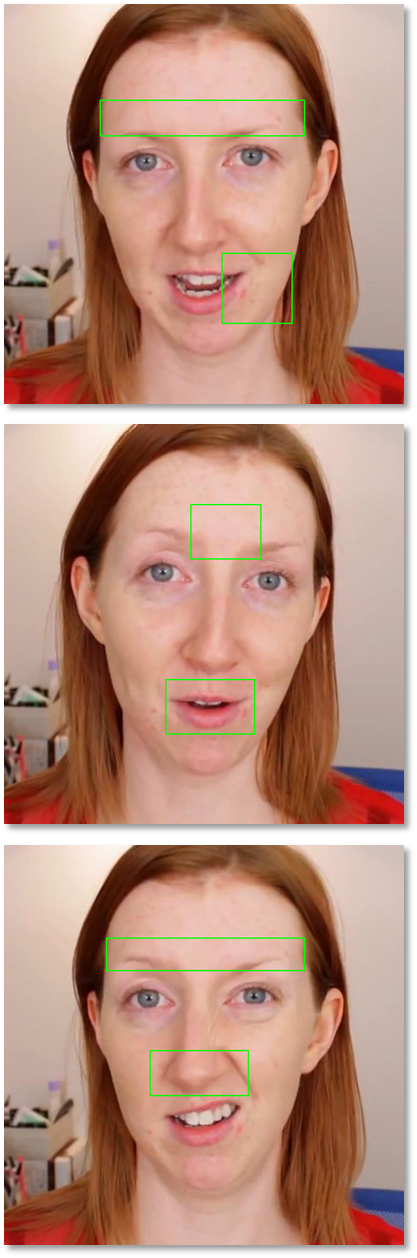}}\hspace{-2pt}
\subfigure[]{\includegraphics[width=3cm]{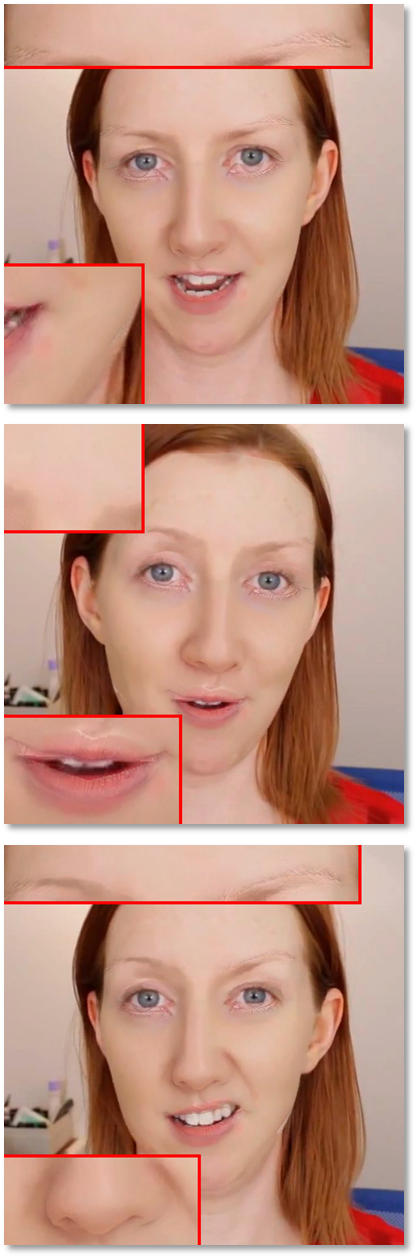}}\hspace{-2pt}
\subfigure[]{\includegraphics[width=3cm]{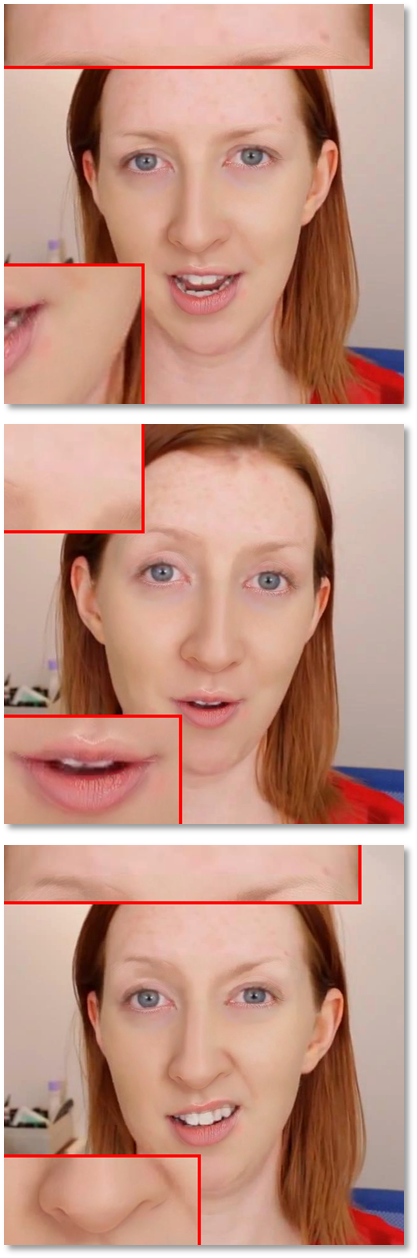}}\hspace{-2pt}
\subfigure[]{\includegraphics[width=3cm]{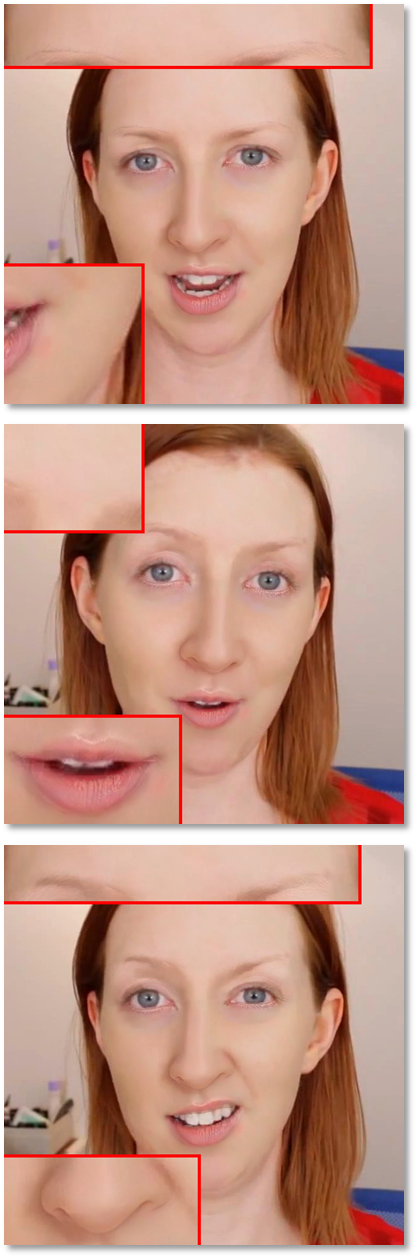}}
\caption{Comparison of makeup transfer results ({\bf Video 6}). (a) Original frames. (b) Result of \cite{Yang:GI:2012}. (c) Our result without forehead points. (d) Our result with forehead points.}
\label{fig:makeup}
\end{figure}

\subsection{Makeup Transfer}
Modifying only the feature landmarks may not significantly improve facial attractiveness, since inherent skin features (e.g., freckles and acnes) may subjectively affect the attractiveness ratings. Inspired by the idea of makeup transfer, the skin detail of a face can also be improved by using another face with better skin detail.

In contrast to \cite{Guo:2009}, we use a template dataset of ten faces with realistic makeup to select a template associated with the skin color that is close to the input face. To improve the forehead skin, we manually add seven feature points on the boundary of forehead region for the first frame. The corresponding forehead points in the subsequent frames are located by adding the difference of $p$ (the position between the eyebrows) between two consecutive frames since the difference is small between frames. According to the feature points of each original frame and the selected template, we can obtain the warped versions of the template. The face regions are approximately aligned. 

\begin{figure}[!t]
\begin{center}
\includegraphics[width=11cm]{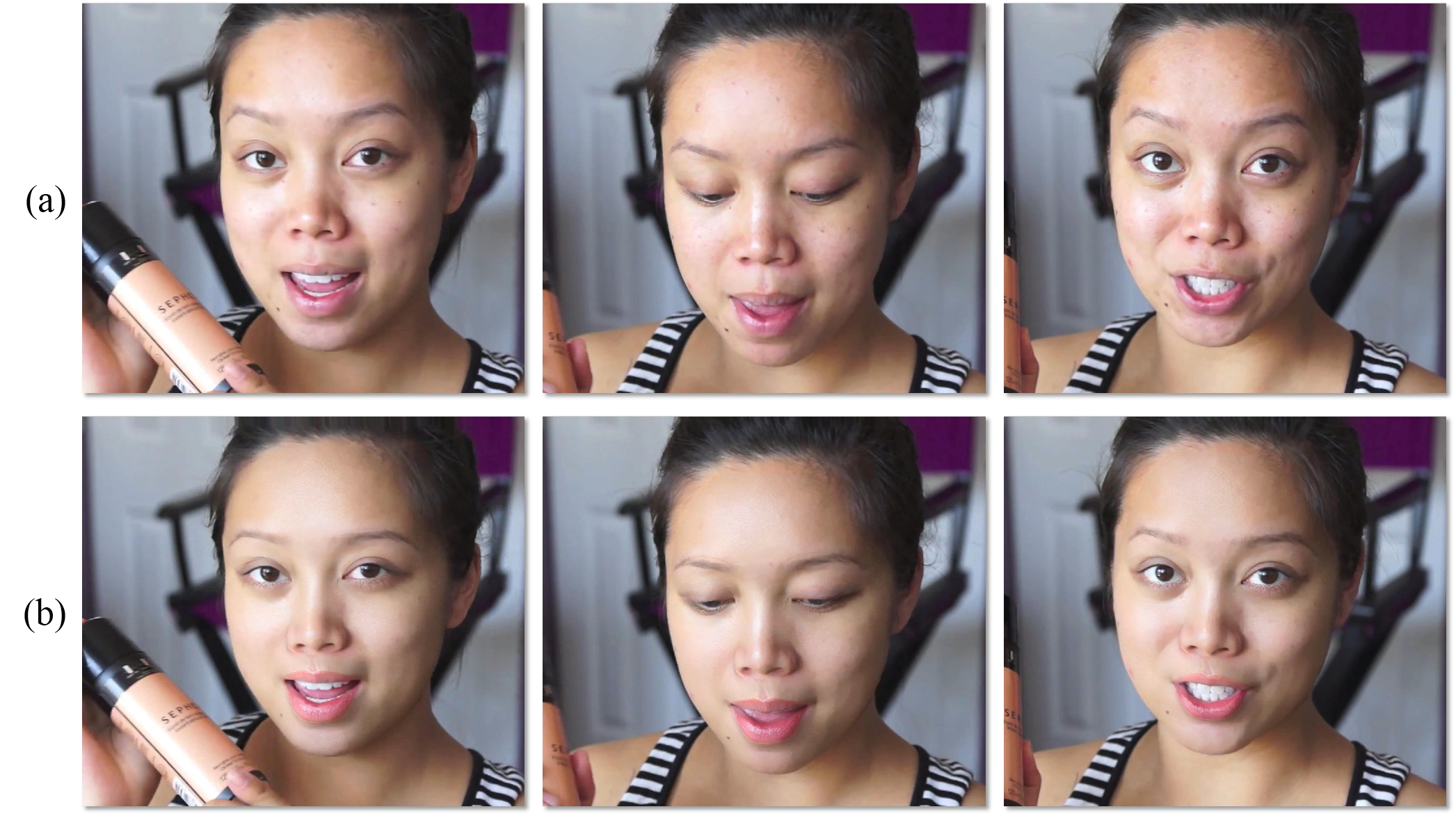}
\caption{Beautified results. (a) Original frames. (b) Our result (facial attractiveness enhancement + makeup transfer).}
\label{fig:beauty}
\end{center}
\end{figure}

All the original frames and the warped templates are then separately decomposed into three layers: a structure layer, a skin layer, and a color layer (refer to the subsections 3.3 and 3.4 in \cite{Guo:2009}). For each original frame, the information in the skin layer is modified with a weighted addition of that of the warped template and itself. The color layer of the warped template is transferred onto that of the corresponding original frame using alpha blending. Then, the intrinsic structure layer, the modified skin layer, and the transferred color layer of each original frame are composed together to obtain the synthesized frame. To make natural-looking of the face boundary, we use Poisson method \cite{Perez:2003} to improve the final result. The comparison of makeup transfer results is shown in Fig.~\ref{fig:makeup}, where the results obtained by the proposed approach associated with manually labelled forehead points are better. Furthermore, we can beautify an input face in a video by simultaneously enhancing facial attractiveness and improving skin details, as shown in Fig.~\ref{fig:beauty}. Results and comparisons are shown in the online videos\footnote{\url{http://youtu.be/xdCF9RyIuOM}}.

\begin{figure}[!ht]
\begin{center}
\includegraphics[width=14cm]{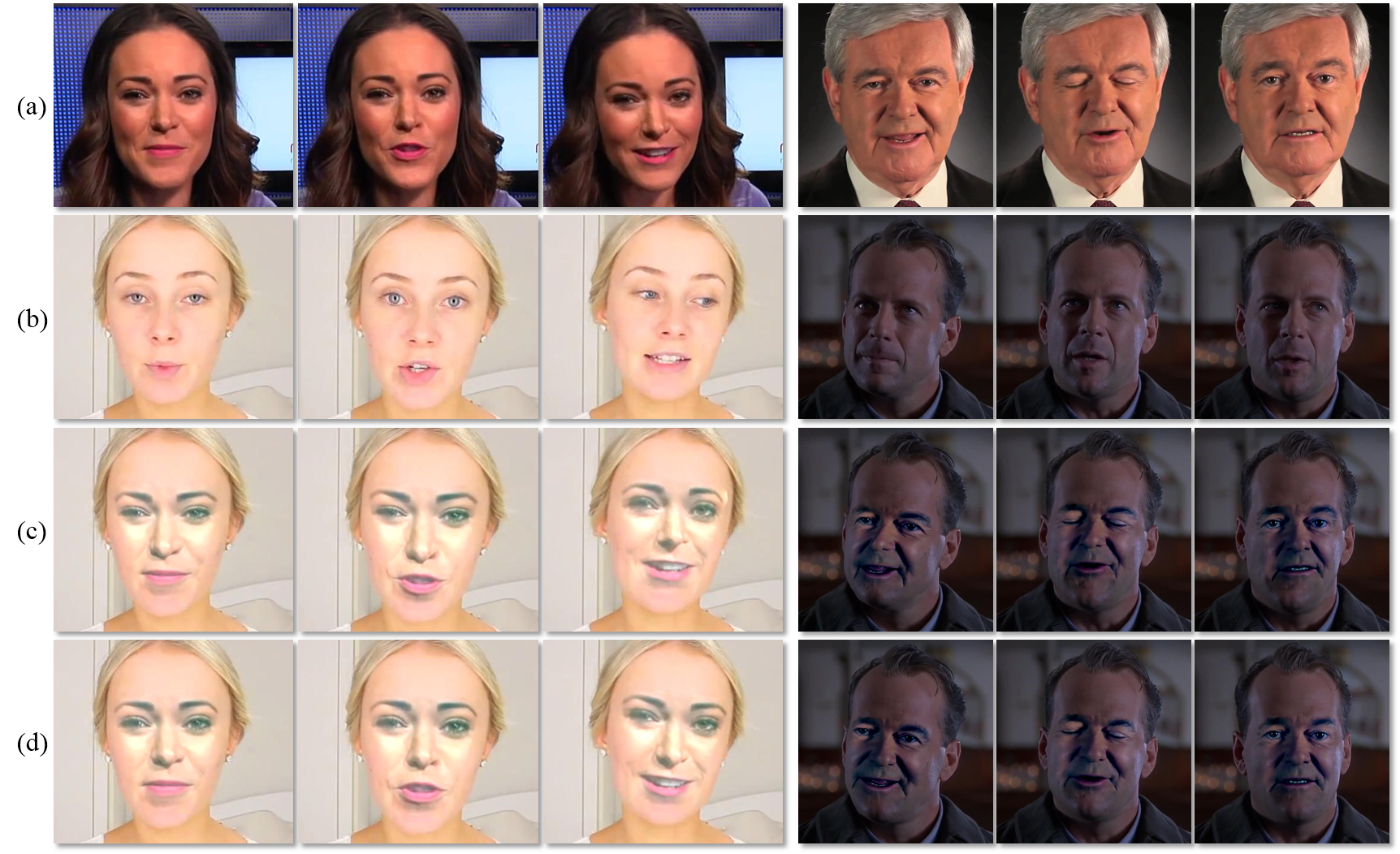}
\caption{Comparison of face replacement results on {\bf Video 8} and {\bf Video 9}. (a) Retimes frames after RCTW. (b) Target frames. (c) Dale's method \cite{Dale:2011}. (d) Our result.}
\label{fig:replace}
\end{center}
\end{figure}

\subsection{Face Replacement}
Given a source video with the desired face and a target video, the facial feature points for each frame are marked using ISDM method. The source frames are retimed using the robust canonical time warping (RCTW) technique \cite{Panagakis:2013:CVPR} for better matching with the expressions of the face in the target video. The facial feature points of each retimed source frame are projected on the corresponding target frame using affine transformation. Based on the feature point pairs, TSSW produces warping surfaces that map the retimed source frames to the projected points. The warped frames and the target frames are aligned followed by seam computation. To further create a truly photo-realistic composite, we apply the Poisson method to blend the face boundary.

Compared to \cite{Dale:2011}, we do not use a 3D-tensor face model, which often relies on intensive user interactions in some key-frames to track the attribute parameters. Only 2D facial feature points are required in our face editing system, which reduces manual costs. In addition, we retime the source frames rather than the target frames, which has the advantage in ensuring that the expressions of the retimed source frames matches the subtitles or voice in the target video. The comparison of synthesized results is shown in Fig.~\ref{fig:replace} and the online video\footnote{\url{http://youtu.be/ZL1hncJ9BMA}}.

\begin{figure}[!t]
\begin{center}
\includegraphics[width=14cm]{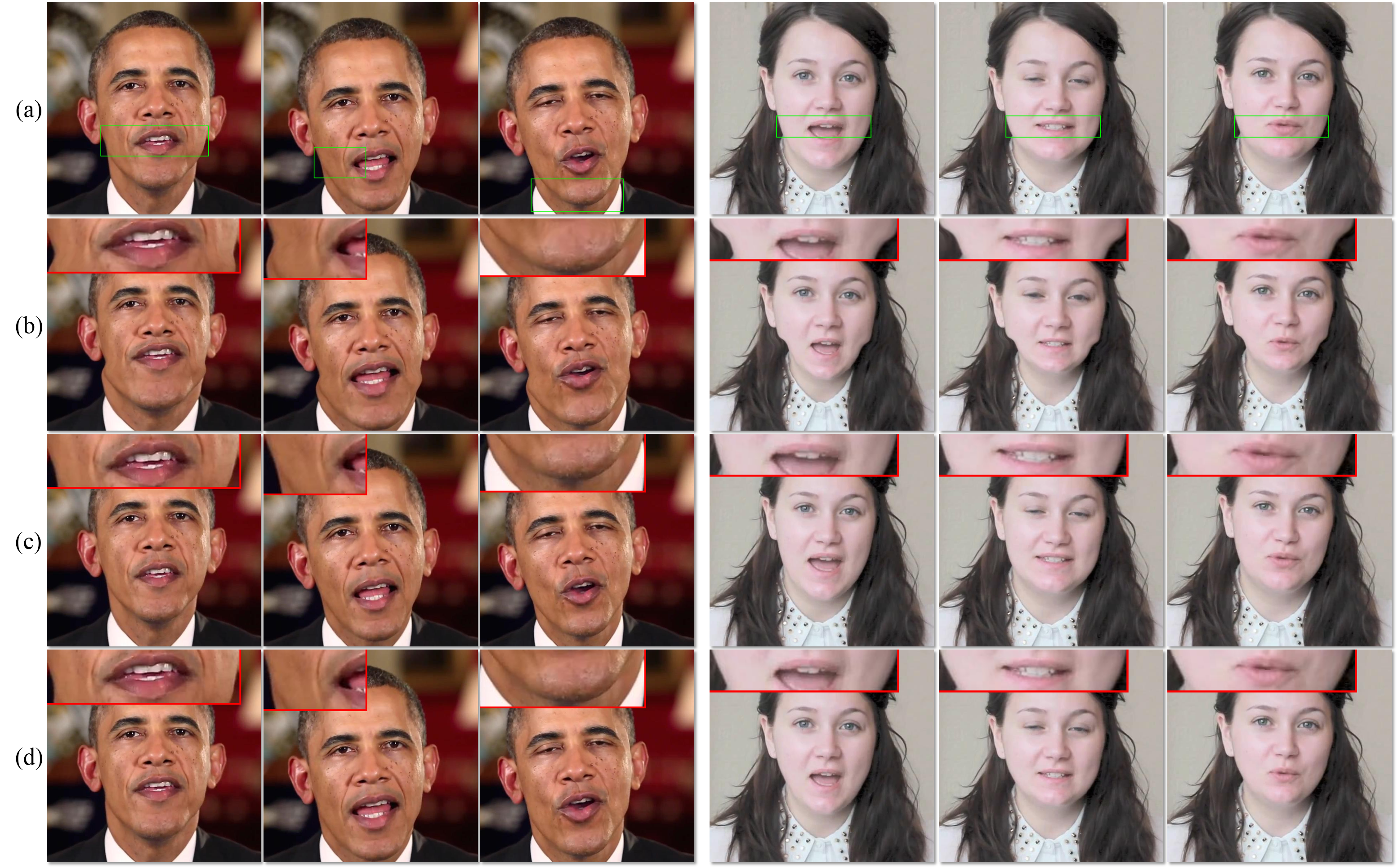}
\caption{Comparison of expression maipulation results. Left with manipulation factor ${\it c}=1.15$ ({\bf Video 10}) and right with ${\it c}=0.85$ ({\bf Video 11}). (a) Original frames. (b) Result of \cite{Ma:MRLS:2013}. (c) Result of \cite{Yang:2012}. (d) Our result.}
\label{fig:express}
\end{center}
\end{figure}

\subsection{Expression Manipulation}
To manipulate (e.g., exaggerate or neutralize) facial expressions across the video sequence while preserving the identity of the input face and 2D pose information of the head, the positions of facial feature points need to be adjusted (especially in the mouth region) with a fixed manipulation factor ${\it c}$, and then projected on each frame for constructing feature point pairs. Through TSSW, the warped frames can be obtained by the estimated warping surfaces. 

Compared to \cite{Yang:2012}, we do not require fitting a video sequence and a dataset of 3D face models. For example, a person changes his expressions from neutral to smile and then back to neutral expression. Our approach is robust to any expressions in a video for exaggeration (${\it c}>1$) and neutralization (${\it c}<1$). Fig.~\ref{fig:express} and the video demonstration\footnote{\url{http://youtu.be/mhzNP3CF0uM}} show results and comparisons of exaggeration and neutralization, and indicate that our approach produces realistic results (even though sometimes wrong point positions, while \cite{Ma:MRLS:2013} is sensitive to positions) and preserves temporal coherence in the synthesized videos.

\begin{figure}[!t]
\begin{center}
\includegraphics[width=10cm]{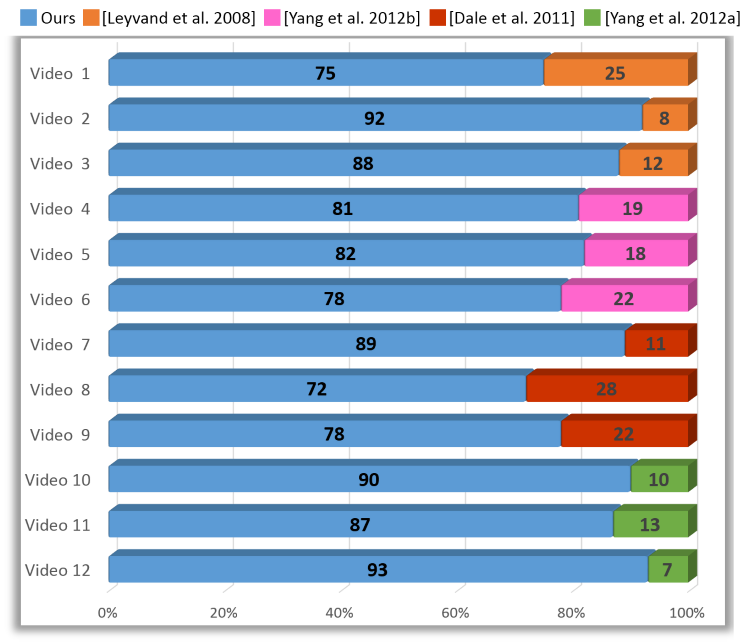}
\caption{The stacked bar chart of participants' preferences for our methods compared with \cite{Leyvand:2008}, \cite{Yang:GI:2012}, \cite{Dale:2011}, and \cite{Yang:2012} among {\bf Video 1 $\sim$ 12}.}
\label{fig:bar}
\end{center}
\end{figure}

\subsection{User Study}
Since without the reference videos, a subjective evaluation obtained by human observers is probably the best way to validate the effectiveness of video face editing. This is due to the sensitivity of human observers to the visual information in the resultant videos. There exist several complicated approaches based on the subjective results, such as \cite{Xu:2008:SIGIR:EFP}, \cite{Nguyen:2011:IJCAI}, \cite{Reches:2014:ACMTIST:FEC}, and \cite{Li:2014:ACMTIST:LPH}. Followed by \cite{Song:2010:TPAMI:VCP}, we exploit the paired comparison method and perform a user study on {\it Amazon Mechanical Turk}\footnote{https://requester.mturk.com/}, to validate the effectiveness of the proposed approaches. For each video, we invite 100 participants coming from diverse backgrounds and aged 20 to 45 years old. The participants are presented with two synthesized results side by side at one time and then asked to choose they preferred video. During the survey, videos obtained by different methods are randomly ordered to avoid bias. Participants may lose the patience in a long time user study. Therefore, each pair of videos is constructed about 15 $\sim$ 50 seconds at the 12 fps frame rate. 

In this survey, we mainly compared our approaches with the existing face editing algorithms, i.e., results obtained by \cite{Leyvand:2008}, \cite{Yang:GI:2012}, \cite{Dale:2011}, and \cite{Yang:2012} for facial attractiveness enhancement, makeup transfer, face replacement, and expression manipulation applications, respectively. For facial attractiveness enhancement, we ask the participants to select the more attractive face (particularly unchanged face geometry throughout the video) in each pair. For makeup transfer, the participants are asked to indicate the face with better skin details in which side of each pair. For replacement, they are asked to tick the more natural-looking face (especially with similar luminance of face region across the video sequence) in each pair. For expression manipulation, we ask them to select the more stable background in each pair. Fig.~\ref{fig:bar} shows the participants' preference among examples from Video 1 to Video 12, which indicates that qualitative empirical results obtained by our proposed approaches for the above four applications are better than those obtained by existing methods.

\subsection{Reconstruction Accuracy}

\begin{figure}[!t]
\normalsize
\centering
\subfigure[]{\includegraphics[width=3.5cm]{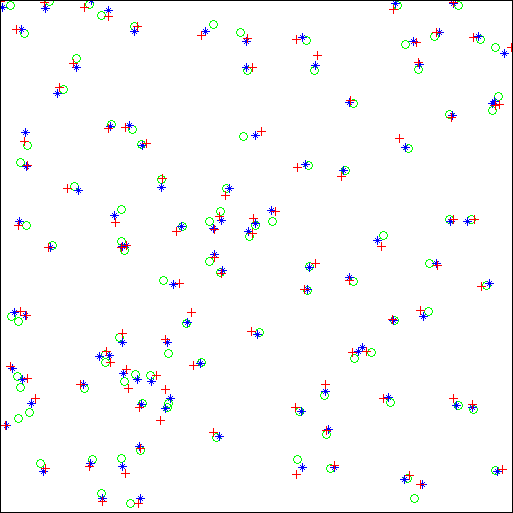}}\hspace{5pt}
\subfigure[]{\includegraphics[width=3.5cm]{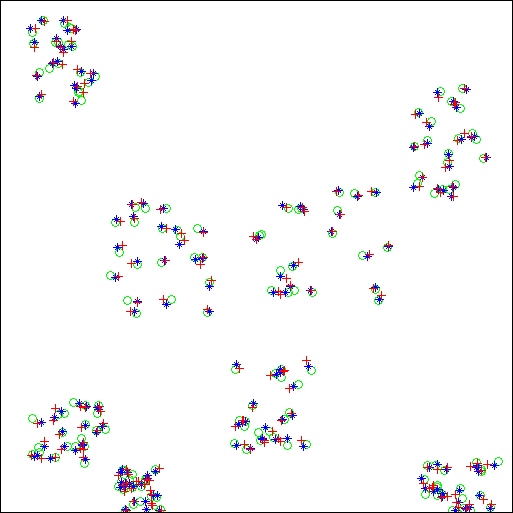}}\hspace{5pt}
\subfigure[]{\includegraphics[width=3.5cm]{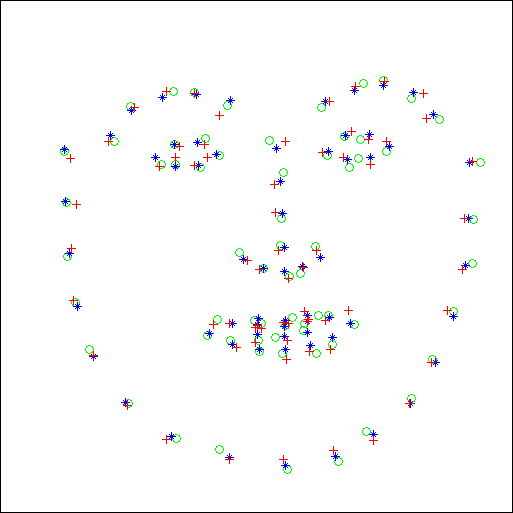}}
\caption{Sampling positions for test functins, where ``o'': $\mathscr{P}_0$; ``$*$'': $\mathscr{P}_1$; ``+'': $\mathscr{P}_2$. (a) R100. (b) C160. (c) F66.}
\label{fig:sampling}
\end{figure}

To demonstrate the reconstruction accuracy of warping surfaces using the existing MBA algorithm and the proposed TSSW approach, we conduct experiments on several test functions. First, we use three types of sampled data points, shown in Fig.~\ref{fig:sampling}. R100 represents 100 points randomly sampled. In C160, the sampled points are divided into 8 clusters, each of which has 20 sampled points. F66 consists of 66 data points sampled from active appearance model (AAM) \cite{Li:2013:ACMTIST:SAM}, which is similar to real-world 2D face geometry. For each data set, the positions marked with ``o'', ``$*$'', and ``+'' symbols represent the feature points in the previous frame, current frame and next frame, respectively, which are denoted as $\mathscr{P}_0$, $\mathscr{P}_1$, and $\mathscr{P}_2$.

We selected five test functions used in \cite{Nielson:CGA:1993}. The resulting test functions are, for $0\leq {\it x, y}\leq1$,

\begin{equation*}
\begin{split}
g_1(x,y) = & 0.75{\rm exp}\left(-(9x+1)^2/49-(9y+1)/10\right)-0.2{\rm exp}\left(-(9x-4)^2-(9y-7)^2\right).\\ \vspace{10pt}
g_2(x,y) =& \left({\rm tanh}(9-9x-9y)+1\right)/9.\\ 
g_3(x,y) =& \left(1.25+{\rm cos}(5.4y)\right)/\left(6+6(3x-1)^2\right).\\ 
g_4(x,y) =& {\rm exp}\left(-20.25(x-0.5)^2-20.25(y-0.5)^2\right)/3.\\
g_5(x,y) =& \sqrt{64/81-(x-0.5)^2-(y-0.5)^2}-0.5.
\end{split}
\end{equation*}

For each test function ${\it g_r} (1\leq {\it r}\leq 5)$, we compute the horizontal and vertical warping surfaces by applying MBA and TSSW to the above three data sets (i.e., R100, C160, and F66). For simplicity, we denote ${\it g_{r,k}}={\it g_r(x_{s,k},y_{s,k})}$, where ${\it x_{s,k}=(x_k-{\rm1})/(M-{\rm1})}$, ${\it y_{s,k}=(y_k-{\rm1})/(N-{\rm1})}$, for $1\leq{\it x_k}\leq{\it M}$, $1\leq{\it y_k}\leq{\it N}$. To evaluate the reconstruction accuracy, we choose ${\it M=N=}512$. The root mean square (RMS) error between the test function ${\it g_r}$ and the approximation function ${\it f({\rm \Psi}_t^l)}$ can be measured on the feature point positions $\mathscr{P}_1$, and $\mathscr{P}_2$ as follows.

\begin{equation}
\begin{split}
RMS=\sqrt{\sum_{l=1}^2\sum_{t=1}^T\sum_{(x_k,y_k)\in \mathscr{P}_t}\left(g_{r,k}-f_k({\rm\Psi}_t^l)\right)^2/KT},
\end{split}
\end{equation}
where ${\it T}=2$, and for R100, ${\it K}=100$; for C160, ${\it K}=160$; for F66, ${\it K}=66$.  Table~\ref{lab:rmse} shows the quantitative results, which demonstrate that the proposed TSSW approach achieves better reconstruction accuracy than the conventional MBA algorithm regardless of the type of the test function.

\begin{table}[!ht]
\begin{center}
\caption{Comparison of ${\it RMS}$ Errors of five test functions}{
\begin{tabular}{|c|c|c|c|c|c|c|}
\hline
\multirow{2}{*}{Functions} &
\multicolumn{3}{c|}{MBA} &
\multicolumn{3}{c|}{TSSW} \\
\cline{2-7} & R100 & C160 & F66 & R100 & C160 & F66 \\
\hline
${\it g}_1$ & 5.000 &	2.467&	5.188&		4.959&	2.155&	4.666\\
\hline
${\it g}_2$ & 5.032 & 2.466 & 5.197 &	4.978 &	2.153 & 4.659 \\
\hline
${\it g}_3$ & 5.028	&2.451&	5.183&		4.968&	2.189&	4.644\\
\hline
${\it g}_4$ & 5.044&	2.469&	5.172&		4.991&	2.168&	4.703\\
\hline
${\it g}_5$ &5.002&	2.458&	5.179&		4.979&	2.170&	4.743\\
\hline
\end{tabular} }
\label{lab:rmse}
\end{center}
\end{table}

\section{Conclusions and Future Work}
In this paper we have developed a novel, video-based warping method, TSSW, for face editing using an energy function containing a data-driven term, a smoothness term, and feature point constraints. TSSW has been successfully applied to four example tasks (facial attractiveness enhancement, makeup transfer, face replacement, and expression manipulation), which also has the potential for use in facial expression synthesis and facial image dubbing in videos, where temporal-spatial coherence is also required to maintain. Notably, each control lattice can be solved using its corresponding Euler-Lagrange equation. One major advantage of our approach is that it allows natural editing of a face in a video even when there is a complicated background. Moreover, this approach does not require user interactions and therefore significantly saves manual costs.
\newline \indent {\it Limitations.} First, our algorithm uses the improved version of SDM to achieve facial feature localization, it may be difficult to obtain good initialization of tracking for all the input videos. Second, due to the lack of 3D information, our method is suboptimal for large pose variations where complex facial geometry and the dynamic elements of faces need to be synthesized. In practice, the method performs well as long as the pose differences between two consecutive frames are not very large. 

In the future we plan to extend TSSW with the data-driven enhancement of face editing for general pose. Furthermore, another future work is to improve the efficiency of control lattice estimation in this approach and explore how to apply it to the real-time environment.


\small
\bibliographystyle{model1-num-names}
\bibliography{references}


\end{document}